\documentclass[10pt, a4paper]{article}
\usepackage{lrec}
\usepackage{graphicx}
\usepackage{tabularx}
\usepackage{soul}

\usepackage{epstopdf}
\usepackage[latin1]{inputenc}

\usepackage{hyperref}
\usepackage{xstring}

\usepackage{amsmath}
\usepackage{amssymb}
\usepackage{subfigure}
\usepackage{booktabs}
\usepackage{array}
\usepackage{multirow}
\usepackage[export]{adjustbox}
\newcolumntype{L}[1]{>{\raggedright\let\newline\\\arraybackslash\hspace{0pt}}m{#1}}
\newcolumntype{C}[1]{>{\centering\let\newline\\\arraybackslash\hspace{0pt}}m{#1}}
\newcolumntype{R}[1]{>{\raggedleft\let\newline\\\arraybackslash\hspace{0pt}}m{#1}}

\newcommand\Tstrut{\rule{0pt}{2.5ex}}         

\title{Propagate-Selector: Detecting Supporting Sentences for \\ Question Answering via Graph Neural Networks}

\name{Seunghyun Yoon$^{1}$, Franck Dernoncourt$^{2}$, Doo Soon Kim$^{2}$, Trung Bui$^{2}$, Kyomin Jung$^{1}$}

\address{
$^{1}$Department of Electrical and Computer Engineering, Seoul National University, Seoul, Korea \\
$^{2}$Adobe Research, San Jose, CA, USA\\
\{mysmilesh, kjung\}@snu.ac.kr, \{franck.dernoncourt, dkim, bui\}@adobe.com\\
}

\abstract{
In this study, we propose a novel graph neural network called propagate-selector (PS), which propagates information over sentences to understand information that cannot be inferred when considering sentences in isolation.
First, we design a graph structure in which each node represents an individual sentence, and some pairs of nodes are selectively connected based on the text structure.
Then, we develop an iterative attentive aggregation and a skip-combine method in which a node interacts with its neighborhood nodes to accumulate the necessary information.
To evaluate the performance of the proposed approaches, we conduct experiments with the standard HotpotQA dataset. The empirical results demonstrate the superiority of our proposed approach, which obtains the best performances, compared to the widely used answer-selection models that do not consider the intersentential relationship.
\\ \newline \Keywords{question answering, supporting sentence, graph neural network} }

\begin{document}

\maketitleabstract

\section{Introduction}
\label{introduction}
Understanding texts and being able to answer a question posed by a human is a long-standing goal in the artificial intelligence field. With the rapid advancement of neural network-based models and the availability of large-scale datasets, such as SQuAD~\cite{rajpurkar2016squad} and TriviaQA~\cite{joshi2017triviaqa}, researchers have begun to concentrate on building automatic question-answering (QA) systems. One example of such a system is the machine-reading question-answering (MRQA) model, which provides answers to questions from given passages~\cite{seo2016bidirectional,xiong2016dynamic,wang2017gated,shen2017reasonet}.

Recently, research has revealed that most questions in existing MRQA datasets do not require reasoning across sentences in the given context (passage); instead, they can be answered by looking at only a single sentence~\cite{weissenborn2017making}.
Using this characteristic, a simple model can achieve performance competitive with that of a sophisticated model.
However, in most real scenarios of QA applications, more than one sentence should be utilized to extract a correct answer.

\begin{figure}[t]
\centering
\includegraphics[width=1.0\columnwidth]{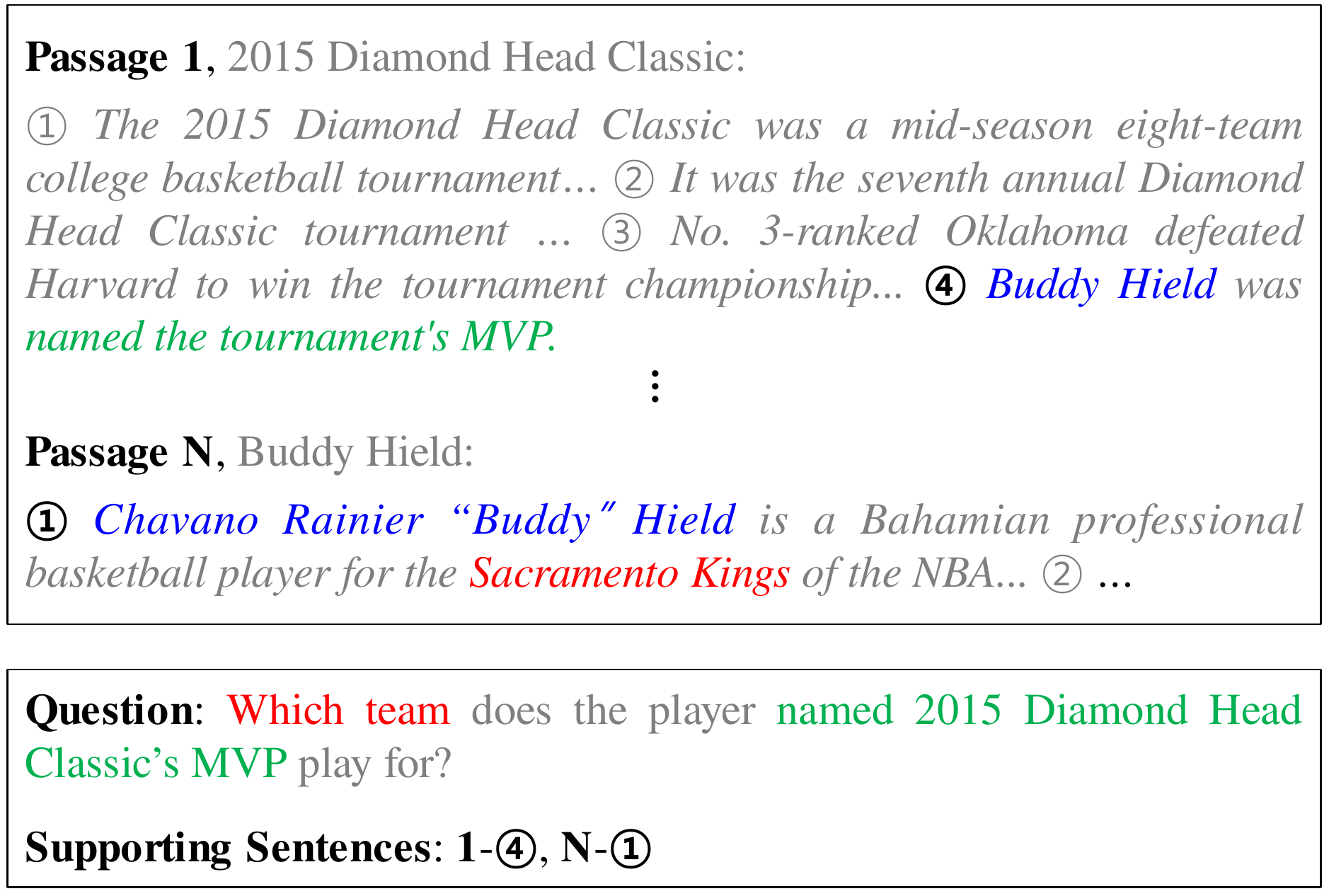}
\caption{
An example of dataset. Detecting \textit{supporting sentences} is an essential step being able to answer the question.
}
\label{fig:example}
\end{figure}

To alleviate this limitation of previous datasets, another type of dataset was developed in which answering the question requires reasoning over multiple sentences in the given passages~\cite{yang2018hotpotqa,welbl2018constructing}. 
Figure~\ref{fig:example} shows an example of a recently released dataset, the \textsf{HotpotQA}.
This dataset consists of not only question-answer pairs with context passages but also \textit{supporting sentence} information for answering the question annotated by a human.

In this study, we build a model that exploits the relational information among sentences in passages to classify the \textit{supporting sentences} that contain the essential information for answering the question.
To this end, we propose a novel graph neural network model named \textbf{propagate-selector (PS)}, which can be directly employed as a subsystem in the QA pipeline. First, we design a graph structure to hold information in the \textsf{HotpotQA} dataset by assigning each sentence to an independent graph node. Then, we connect the undirected edges between nodes using a proposed graph topology~(see the discussion in the~\ref{sec_p3:propsoed_method}). Next, we allow \textbf{PS} to propagate information between the nodes through iterative hops to perform reasoning across the given sentences. Through the propagation process, the model learns to understand information that cannot be inferred when considering sentences in isolation.

Unlike the previous studies, this work does not use the exact ``answer span" information while detecting the supporting sentences. It shows a different way of using the \textsf{HotPotQA} dataset and provides researchers new opportunities to develop a subsystem that is integrated into the full-QA systems (i.e., MRQA).
Through experiments, we demonstrate that compared with the widely used answer-selection models~\cite{wang2016compare,bian2017compare,shen2017inter,tran2018context,yoon2019compare}, the proposed method achieves better performance when classifying \textit{supporting sentences}.

\section{Related Work}
\label{sec:realated_work}
Previous researchers have also investigated neural network-based models for MRQA. One line of inquiry employs an attention mechanism between tokens in the question and passage to compute the answer span from the given text~\cite{seo2016bidirectional,wang2017gated}. As the task scope was extended from specific- to open-domain QA, several models have been proposed to select a relevant paragraph from the text to predict the answer span~\cite{wang2018r,clark2018simple}. However, none of these methods have addressed reasoning over multiple sentences.

To understand the relational patterns in the dataset, researchers have also proposed graph neural network algorithms.
\cite{kipf2017semi} proposed a graph convolutional network to classify graph-structured data. This model was further investigated for applications involving large-scale graphs~\cite{hamilton2017inductive}, for the effectiveness of aggregating and combining graph nodes by employing an attention mechanism~\cite{velivckovic2017graph}, and for adopting recurrent node updates~\cite{palm2018recurrent}. 
These methods successfully demonstrated their potential and effectiveness in understanding relational datasets, such as entity linking in heterogeneous knowledge graphs, product recommendation systems, and detecting side effects in drug~\cite{wu2019relation,fan2019graph,zitnik2018modeling}.
In addition, one trial involved applying graph neural networks to QA tasks; however, this usage was limited to the entity level rather than sentence-level understanding~\cite{de2019question}.

\section{Task and Dataset}
\label{sec:problem_and_dataset}
The specific problem we aim to tackle in this study is to classify \textit{supporting sentences} in the MRQA task. We consider the target dataset \textsf{HotpotQA}, by~\cite{yang2018hotpotqa}, which comprises tuples ($<$\textit{Q}, $P_n$, $Y_i$, \textit{A}$>$) in which \textit{Q} is the question, $P_n$ is the set of passages as the given context, and each passage $P\,{\in}\,P_n$ further comprises a set of sentences $S_i$ ($S_i\,{\in}\,P_n)$. Here, $Y_i$ is a binary label indicating whether $S_i$ contains the information required to answer the question, and \textit{A} is the answer. In particular, we call a sentence, $S_s\,{\in}\,S_i$, a \textit{supporting sentence} when $Y_s$ is \textit{true}.
Figure~\ref{fig:example} shows an example of the \textsf{HotpotQA} dataset.

In this study, we do not use the answer information from the dataset; we use only the subsequent tuples $<$\textit{Q}, $P_n$, $Y_i$$>$ when classifying \textit{supporting sentences}.
We believe that this subproblem plays an important role in building a full QA pipeline because the proposed models for this task will be combined with other MRQA models in an end-to-end training process.

\section{Methodology}
\label{sec:methods}
Our objective in this study is to identify \textit{supporting sentences}, among sentences in the given text that contain information essential for answering the question.
To tackle this problem, we first introduce answer-selection models, which are widely studied in the research community.
These models are considered strong baselines since they can be directly applied to our task with the same objective function.
Then we describe our proposed method.

\subsection{Baseline approaches}
\label{sec:baseline_approaches}
We introduce baseline models for the answer-selection task, which have been extensively studied and have proved their efficacy to the research community. 
These models are developed to compute the matching similarity between any pairs of text (the question and the target sentence in our case).

\subsubsection{Compare Aggregate Framework (CompAggr).}
\label{ssec:compaggr}
This model~\cite{wang2016compare} computes the matching similarity between two texts (the question and the target sentence). It consists of attention, comparison, and aggregation parts.


\vspace*{1mm}
\noindent\textbf{Attention: }
The soft alignment of the question $\textbf{Q}\,{\in}\,\mathbb{R}^{d\times Q}$ and target sentence $\textbf{S}\,{\in}\,\mathbb{R}^{d\times S}$ (where $d$ is a dimensionality of word embedding and \textit{Q} and \textit{S} are the length of the sequences in the question and sentence, respectively) is computed by applying an attention mechanism over the column vector in $\textbf{Q}$ for each column vector in $\textbf{S}$.
With the computed alignment, we obtain a corresponding vector $\textbf{A}^Q\,{\in}\,\mathbb{R}^{d\times S}$ as follows:
\begin{equation}
    \begin{aligned}
    & \textbf{A}^Q=\textbf{Q}\,{\cdot}\,\text{softmax}({(\textbf{W}{\textbf{Q}})}^{\intercal}\textbf{S}),\\
    \end{aligned}
    \label{eq:comp_attention}
\end{equation}
where $\textbf{W}$ is a learned model parameter matrix.

\vspace*{1mm}
\noindent\textbf{Comparison: }
An element-wise multiplication is employed as a comparison function to combine each pair of $\textbf{A}^Q$ and $\textbf{S}$ into a vector~$\textbf{C}\,{\in}\,\mathbb{R}^{d\times S}$.

\vspace*{1mm}
\noindent\textbf{Aggregation: }
Kim \shortcite{kim2014convolutional}'s CNN with \textit{n}-types of filters is applied to aggregate all information in the vector \textbf{C}.
Finally, the model employs a fully connected layer to compute the matching score between the question and the target sentence as follows:
\begin{equation}
    \begin{aligned}
    & \textbf{R}=\text{CNN}(\textbf{C}),~~(\textbf{R}\,{\in}\,\mathbb{R}^{nd}), \\
    & \hat{y}_{c} = \text{softmax}((\textbf{R})^\intercal~\textbf{W}+\textbf{b}~), \\
    \end{aligned}
    \label{eq:comp_aggregation}
\end{equation}
where $\hat{y}_{c}$ is the predicted probability for the target class, \textit{c}, and $\textbf{W}\,{\in}\,\mathbb{R}^{nd\times c}$ and bias \textbf{b} are learned model parameters.

The loss function for the model is cross-entropy between predicted labels and true-labels as follows:
\begin{equation}
    \begin{aligned}
    \mathcal{L} = -\log \sum_{i=1}^{N} \sum_{c=1}^{C} y_{i,c} \text{log} (\hat{y}_{i,c}), 
    \end{aligned}
    \label{eq:comp_loss}
\end{equation}
where $y_{i,c}$ is the true label vector and $\hat{y}_{i,c}$ is the predicted probability from the softmax layer.
$C$ is the total number of classes (true and false for this task), and $N$ is the total number of samples used in training.

\subsubsection{CompAggr-kMax.}
This model~\cite{bian2017compare} is an extension of the \textbf{CompAggr} model. The only differences lie in applying attention (in equation~(\ref{eq:comp_attention})) to both the \textbf{Q} and \textbf{S} side and applying k-max pooling before the softmax function as follows:
    \begin{equation}
    \begin{aligned}
    & \textbf{A}^Q=\textbf{Q}\,{\cdot}\,\text{softmax}(\text{kMax}({(\textbf{W}{\textbf{Q}})}^{\intercal}\textbf{S})),\\
    & \textbf{A}^S=\textbf{S}\,{\cdot}\,\text{softmax}(\text{kMax}({(\textbf{W}{\textbf{S}})}^{\intercal}\textbf{Q})).\\
    \end{aligned}
    \label{eq:compclip_attention}
\end{equation}

\subsubsection{CompClip-LM-LC.} 
This model~\cite{yoon2019compare} is an extension of the \textbf{CompAggr-kMax} model. It employs the ELMo~\cite{peters2018deep} model to enhance the word embedding layer for the question and target sentence by adopting the pretrained contextual language model. Additionally, it develops a latent clustering method to compute topic information in texts automatically and to use it as auxiliary information to improve the model performance.

\subsubsection{IWAN.}
This model~\cite{shen2017inter} is a variation model based on the compare aggregate framework. Unlike \textbf{CompAggr}, it employs RNNs to encode a sequence of the words in the text (question and target sentence independently).
At the same time, it computes an inter-alignment weight between the question and the target sentence. 
The matching score is computed by aggregating this information (question, target sentence, and inter-aligned representation).

\subsubsection{sCARNN.}
This model~\cite{tran2018context} is an extension of the \textbf{IWAN} model.
It proposes a novel recurrent unit to regulate the flow of the input (sequence of words in a text) and then replaces the RNNs in the \textbf{IWAN} model of the proposed unit.

\subsection{Propagate-Selector}
\label{sec_p3:propsoed_method}
\begin{figure}[t]
\centering
\includegraphics[width=0.95\columnwidth]{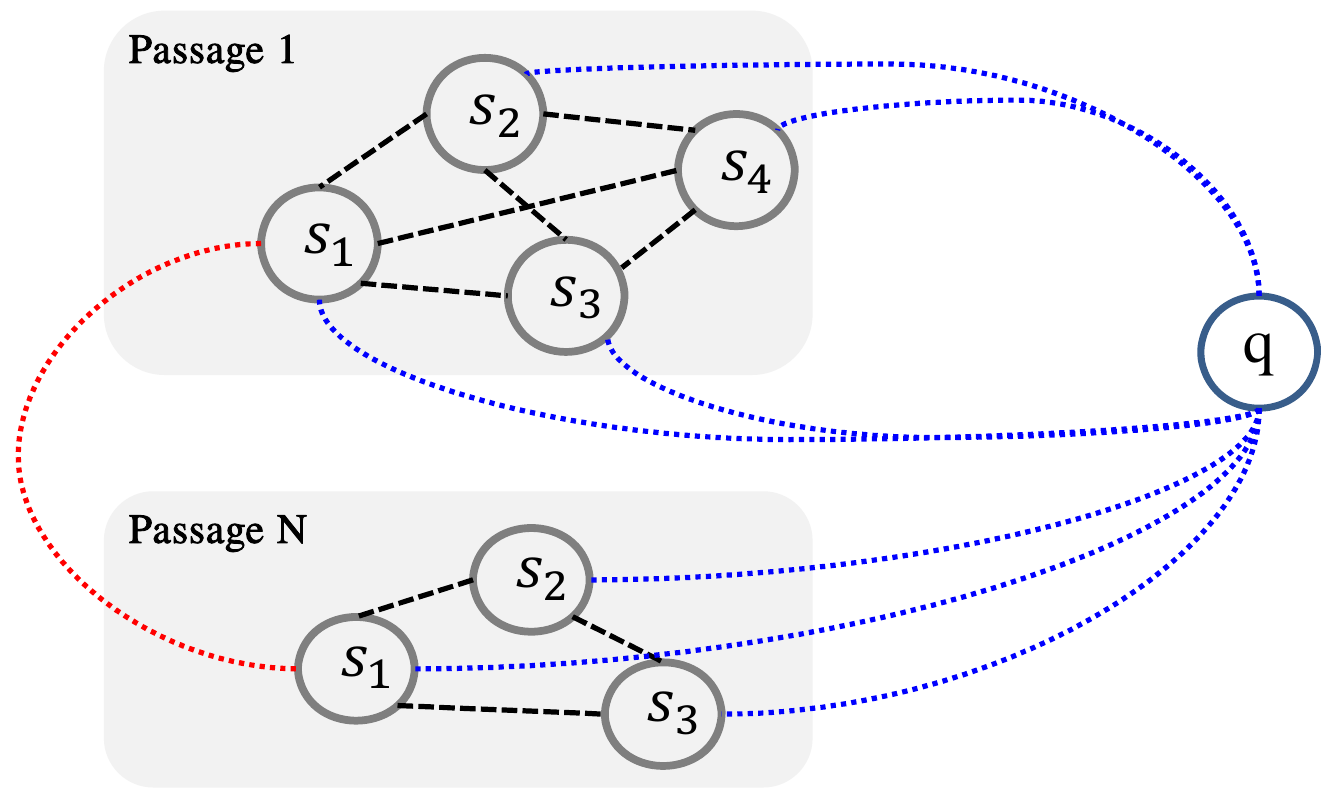}
\caption{
Topology of the proposed model. Each node represents a sentence from the passage and the question. 
}
\label{fig:topology}
\end{figure}

To build a model that can perform reasoning across multiple sentences, we propose a graph neural network model called \textbf{Propagate-selector} (\textbf{PS}). \textbf{PS} consists of four parts as follows (topology, node representation, aggregation, and update):

\vspace*{1mm}
\noindent\textbf{Topology: }
The topology of the graph determines the connections among the nodes in the graph.
These connections will be used as a path that allows the information to flow from one node to another.
Figure~\ref{fig:topology} depicts the topology of the proposed model.
In an offline step, we organize the content of each instance in a graph, where each node represents a sentence from the passages and the question.
Then, we add edges between nodes using the following topology:
\begin{itemize}
\item we fully connect nodes that represent sentences from the same passage (dotted-black);
\item we fully connect nodes that represent the first sentence of each passage (dotted-red);
\item we add an edge between the question and every node for each passage (dotted-blue).
\end{itemize}
In this way, we enable a path by which sentence nodes can propagate information between both inner and outer passages.
Furthermore, we investigate different strategies for connecting those nodes in the graph and examine their corresponding effectiveness for detecting \textit{supporting sentences} in the text (see the discussion in section~\ref{ssec:topology_analysis}).

\vspace*{1mm}
\noindent\textbf{Node representation: }
Question $\textbf{Q}\,{\in}\,\mathbb{R}^{d\times Q}$ and sentence ${\textbf{S}}_i\,{\in}\,\mathbb{R}^{d\times S_i}$ (where $d$ is the dimensionality of the word embedding and $Q$ and ${S}_i$ represent the lengths of the sequences in \textbf{Q} and ${\textbf{S}}_i$, respectively) are processed to acquire the sentence-level information.
Recent studies have shown that a pretrained language model helps the model capture the contextual meaning of words in the sentence~\cite{peters2018deep,devlin2019bert}.
Following this study, we select an ELMo~\cite{peters2018deep} language model for the word-embedding layer of our model as follows:

\begin{equation}
    \begin{aligned}
    &\textbf{L}^Q=\text{ELMo}(\textbf{Q}),~(\textbf{L}^Q{\in}\mathbb{R}^{d\times Q}),\\
    &\textbf{L}^S=\text{ELMo}(\textbf{S}),~(\textbf{L}^S{\in}\mathbb{R}^{d\times S}).\\
    \end{aligned}
    \label{eq:elmo}
\end{equation}

Using these new representations, we compute the sentence representation as follows:
\begin{equation}
    \begin{aligned}
    & \textbf{h}^Q_t = f_{\theta}(\textbf{h}^Q_{t-1},\textbf{L}^Q_t), \\
    & \textbf{h}^{S}_t=f_{\theta}(\textbf{h}^{S}_{t-1},\textbf{L}^S_t), \\
    & \textbf{N}^Q=\textbf{h}^Q_{\text{last}},~~\textbf{N}^S=\textbf{h}^S_{\text{last}},
    \end{aligned}
    \label{eq:node_representation}
\end{equation}
where $f_\theta$ is the RNN function with the weight parameters $\theta$ and ~$\textbf{N}^Q\,{\in}\,\mathbb{R}^{d'}$ and~$\textbf{N}^S\,{\in}\,\mathbb{R}^{d'}$ are node representations for the question and sentence, respectively (where $d'$ is the dimensionality of the RNN hidden units).

As computing the node representation is an essential process for acquiring information from a text, we investigate various approaches for encoding sentences, such as replacing the ELMo word representations using different methods (the GloVe~\cite{pennington2014glove} or the BERT~\cite{devlin2019bert}) and replacing the RNN function in equation~(\ref{eq:node_representation}) with the pooling method.
Furthermore, we adopt the universal sentence encoding method based on the recently developed transformer model~\cite{cer2018universal}. Detailed information will be given in the section~\ref{ssec:node_analysis}

\vspace*{1mm}
\noindent\textbf{Aggregation: }
An iterative attentive aggregation function to the neighbor nodes is utilized to compute the amount of information to be propagated to each node in the graph as follows:

\begin{equation}
\begin{aligned}
& \textbf{A}_v^{(k)}=\sigma({\sum_{u\in N(v)}} ~a_{vu}^{(k)}~\textbf{W}^{(k)}\cdot~\textbf{N}_u^{(k)}), \\
& a_{vu}^{(k)}=\frac{\text{exp}(\textbf{S}_{vu})}{{\sum_k}\text{exp}(\textbf{S}_{vk})}, \\
& \textbf{S}_{vu}^{(k)}=(\textbf{N}_v^{(k)})^\intercal\cdot\textbf{W}^{(k)}\cdot~{\textbf{N}_u^{(k)}},
\end{aligned}
\label{eq:aggregation}
\end{equation}
where $\textbf{A}_v\,{\in}\,\mathbb{R}^{d'}$ is the aggregated information for the \textit{v}-th node computed by attentive weighted summation of its neighbor nodes, $a_{vu}$ is the attention weight between node \textit{v} and its neighbor nodes $u~(u{\in}N(v))$, $\textbf{N}_u\,{\in}\,\mathbb{R}^{d'}$ is the \textit{u}-th node representation, $\sigma$ is a nonlinear activation function, and $\textbf{W}\,{\in}\,\mathbb{R}^{d'\times d'}$ is the learned model parameter. Because all the nodes belong to a graph structure in which the iterative aggregation is performed among nodes, the \textit{k} in the equation indicates that the computation occurs in the \textit{k}-th hop (iteration).

\vspace*{1mm}
\noindent\textbf{Update: }
The aggregated information for the \textit{v}-th node, $\textbf{A}_v$ in equation~(\ref{eq:aggregation}), is combined with its previous node representation to update the node. We apply a skip connection to allow the model to learn the amount of information to be updated in each hop as follows:

\begin{equation}
\begin{aligned}
& \textbf{N}_v^{(k)}=\sigma(\textbf{W}'\cdot\{\textbf{N}_v^{(k-1)};\textbf{A}_v^{(k)}\}), \\
\end{aligned}
\label{eq:combine}
\end{equation}
where $\sigma$ is a nonlinear activation function,~\{;\} indicates vector concatenation, and $\textbf{W}'\,{\in}\,\mathbb{R}^{d'\times 2d'}$ is the learned model parameter.

\subsection{Optimization}
\label{sec:optimization}
Because our objective is to classify \textit{supporting sentences} ($S_i\,{\in}\,{P_n}$) from the given tuples $<$\textit{Q}, $P_n$, $Y_i$$>$, we define two types of loss to be minimized.
One is a rank loss that computes the cross-entropy loss between a question and each sentence using the ground-truth $Y_i$ as follows:

\begin{equation}
\begin{aligned}
& \text{loss}_{rank}=-\text{log}\,{\sum_{i=1}^{N}}~Y_i~\text{log}({S}_i), \\
& \text{S}=[\text{score}_1,...,\text{score}_i], \\
& \text{score}_i=g_{\theta}(\textbf{N}^Q, \textbf{N}^S_i),
\end{aligned}
\label{eq:loss_rank}
\end{equation}
where $g_{\theta}$ is a feedforward network that computes a similarity score between the final representation of the question and each sentence.
The other is attention loss which is defined in each hop, to reward the model when it correctly attends the supporting sentences, as follows:

\begin{equation}
\begin{aligned}
& \text{loss}_{attn}=-\text{log}\,{\sum_{i=1}^{k}}{\sum_{i=1}^{N}}~Y_i~\text{log}(a_{qi}^{(k)}),
\end{aligned}
\label{eq:loss_attention}
\end{equation}
where $a_{qi}^{(k)}$ indicates the relevance between the question node \textit{q} and the \textit{i}-th sentence node in the \textit{k}-th hop as computed by equation~(\ref{eq:aggregation}).

Finally, these two losses are combined to construct the final objective function:

\begin{equation}
\begin{aligned}
& \mathcal{L}=\alpha~\text{loss}_{rank} + \text{loss}_{attn}, 
\end{aligned}
\label{eq:loss_final}
\end{equation}
where $\alpha$ is a hyperparameter.

\begin{table}[t]
\small
\centering
\begin{tabular}{L{0.4\columnwidth}R{0.2\columnwidth}R{0.2\columnwidth}}
\toprule
\textbf{properties} & \textbf{train} & \textbf{dev} \\
\midrule
\# questions & 90,447   & 7,405  \\
\# sentences    & 3,703,344     & 306,487   \\
\midrule
passages / question & 9.95   & 9.95 \\
sentences / passage & 4.12  & 4.16 \\
sentences / question & 40.94  & 41.39 \\
supporting sentences / question & 2.39  & 2.43 \\
\midrule
avg tokens (question) & 17.92   & 15.83 \\
avg tokens (sentence) & 22.38   & 22.41 \\
\bottomrule
\end{tabular}
\caption{
Properties of the dataset
}
\label{table:properties}
\end{table}
\begin{figure*}[t]
\centering
\subfigure[hop-1]{\label{fig_hop_1}\includegraphics[width=0.44\columnwidth]{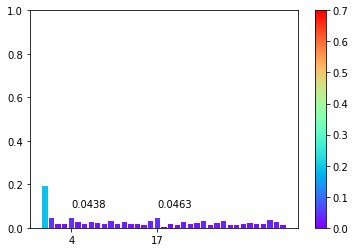}}
\qquad
\subfigure[hop-2]{\label{fig_hop_2}\includegraphics[width=0.44\columnwidth]{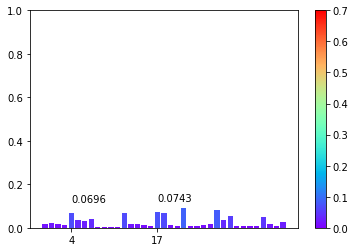}}
\qquad
\subfigure[hop-3]{\label{fig_hop_3}\includegraphics[width=0.44\columnwidth]{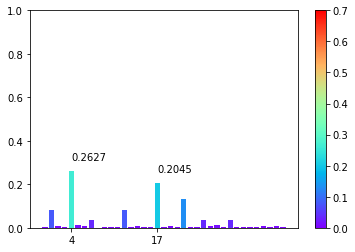}}
\qquad
\subfigure[hop-4]{\label{fig_hop_4}\includegraphics[width=0.44\columnwidth]{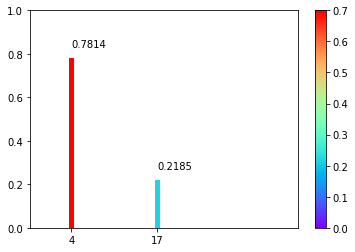}}
\caption{
Attention weights between the question and sentences in the passages. As the number of hops increases, the proposed model correctly classifies \textit{supporting sentences} (ground-truth index 4 and 17). 
}
\label{fig_hop_analysis}
\end{figure*}

\section{Experiments}
\label{sec:experiments}
We regard the task as the problem of selecting the \textit{supporting sentences} from the passages to answer the questions.
Similar to the answer-selection task in the QA literature, we report the model performance using the mean average precision (MAP) and mean reciprocal rank (MRR) metrics. To evaluate the model performance, we use the \textsf{HotpotQA} dataset, which is described in section~\ref{sec:problem_and_dataset}
Table~\ref{table:properties} shows the properties of the dataset.
We conduct a series of experiments to compare baseline methods with the newly proposed models. 
All source code developed to obtain the empirical results will be made available via a public web repository along with the dataset\footnote{http://github.com/david-yoon/propagate-selector}.

\subsection{Implementation Details}
\label{apx:implementation}
To implement the \textbf{propagate-selector} (\textbf{PS}) model, we first use a small version of ELMo (13.6 \textit{M} parameters) that provides 256-dimensional context embedding. This choice was based on the available batch size (50 for our experiments) when training the complete model on a single GPU (GTX 1080 Ti). 
Then, we further experiment with the original version of ELMo (93.6 \textit{M} parameters, 1024-dimensional context embedding). In this case, we were able to increase the batch size only up to 20, which results in excessive training time (approximately 90 hours).
For the sentence encoding, we used a GRU with a hidden unit dimension of 200. 
The hidden unit weight matrix of the GRU is initialized using orthogonal weights~\cite{saxe2013exact}.
Dropout~\cite{srivastava2014dropout} is applied for regularization purposes at a ratio of 0.7 for the GRU (in equation~(\ref{eq:node_representation})) to 0.7 for the attention weight matrix (in equation~(\ref{eq:aggregation})).
For the nonlinear activation function (in equation~(\ref{eq:aggregation}) and (\ref{eq:combine})), we use the $tanh$ function.

Regarding the vocabulary, we replaced vocabulary with fewer than 12 instances in terms of term-frequency with ``\textit{UNK}" tokens. The final vocabulary size was 138,156.
We also applied Adam optimizer~\cite{kingma2014adam}, including gradient clipping by norm at a threshold of 5.

\subsection{Comparisons with Other Methods}
\label{ssec:comparison}
Table~\ref{table:performance_comparison} shows the model performances on the \textsf{HotpotQA} dataset. Because the dataset only provides training (trainset) and validation (devset) subsets, we report the model performances on these datasets. 
While training the model, we implement early termination based on the devset performance and measure the best performance.
To compare the model performances, we choose widely used answer-selection models such as 
\textbf{IWAN}~\cite{shen2017inter}, 
\textbf{sCARNN}~\cite{tran2018context}, 
\textbf{CompAggr}~\cite{wang2016compare}, 
\textbf{CompAggr-kMax}~\cite{bian2017compare}, and \textbf{CompClip-LM-LC}~\cite{yoon2019compare}, which were primarily developed to rank candidate answers for a given question (refer to the section~\ref{sec:baseline_approaches} for detailed information on the models).
In addition to the main proposed model, \textbf{PS}-\textit{rnn-elmo}, we also report the model performance with a small version of ELMo, \textbf{PS}-\textit{rnn-elmo-s}.

As shown in Table~\ref{table:performance_comparison}, the proposed \textbf{PS}-\textit{rnn-elmo} shows a significant MAP performance improvement compared to the previous best model, \textbf{CompClip-LM-LC} (0.702 to 0.734 absolute).

\subsection{Hop Analysis}
\label{ssec:hot_analysis}
Table~\ref{table:performance_hop} shows the model performance (\textbf{PS}-\textit{rnn-elmo}) as the number of hops increases. We find that the model achieves the best performance in the 4-hop case but starts to degrade when the number of hops exceeds 4. We assume that the model experiences the vanishing gradient problem under a larger number of iterative propagations (hops).
Table~\ref{table:performance_hop_small} shows the model performance with the small version of ELMo.

Figure~\ref{fig_hop_analysis} depicts the attention weight between the question node and each sentence node (hop-4 model case). As the hop number increases, we observe that the model properly identifies \textit{supporting sentences} (in this example, sentences \#4 and \#17). This behavior demonstrates that our proposed model correctly learns how to propagate the necessary information among the sentence nodes via the iterative process.

\begin{table}[t]
\small
\centering
\begin{tabular}{L{0.37\columnwidth}C{0.09\columnwidth}C{0.09\columnwidth}C{0.09\columnwidth}C{0.09\columnwidth}}
\toprule
\multicolumn{1}{l}{\multirow{2}{*}{\textbf{Model}\Tstrut}} &  \multicolumn{2}{c}{dev}\Tstrut & \multicolumn{2}{c}{train}\Tstrut \\
\cmidrule{2-5}
\multicolumn{1}{c}{} & MAP\Tstrut & MRR\Tstrut & MAP\Tstrut & MRR\Tstrut \\
\midrule
\textbf{IWAN}~\scriptsize{[1]}\Tstrut     & 0.526\Tstrut & 0.680\Tstrut & 0.605\Tstrut & 0.775\Tstrut \\
\textbf{sCARNN}~\scriptsize{[2]}   & 0.534 & 0.698 & 0.620 & 0.792 \\
\textbf{CompAggr}~\scriptsize{[3]} & 0.659 & 0.812 & 0.796 & 0.911 \\
\textbf{CompAggr-kMax}~\scriptsize{[4]} & 0.670 & 0.825 & 0.767 & 0.901 \\
\textbf{CompClip-LM-LC}~\scriptsize{[5]} & 0.702 & 0.848 & 0.757 & 0.884 \\
\midrule
\textbf{PS}-\textit{rnn-elmo-s}\Tstrut & 0.716\Tstrut & 0.841\Tstrut & 0.813\Tstrut & 0.916\Tstrut \\
\textbf{PS}-\textit{rnn-elmo} & \textbf{0.734} & \textbf{0.853} & \textbf{0.863} & \textbf{0.945} \\
\bottomrule
\end{tabular}
\caption{
Model performance on the HotpotQA dataset (top scores marked in bold).
Models [1-5] are from (Shen et al.,2017a; Tran et al., 2018; Wang and Jiang, 2016; Bian et al.,2017; Yoon et al., 2019), respectively.
}
\label{table:performance_comparison}
\end{table}

\begin{table}[t]
\small
\centering
\begin{tabular}{C{0.27\columnwidth}C{0.11\columnwidth}C{0.11\columnwidth}C{0.11\columnwidth}C{0.11\columnwidth}}
\toprule
\multicolumn{1}{c}{\multirow{2}{*}{\textbf{\# hop}\Tstrut}} & \multicolumn{2}{c}{dev}\Tstrut & \multicolumn{2}{c}{train}\Tstrut \\
\cmidrule{2-5}
\multicolumn{1}{c}{} & MAP\Tstrut & MRR\Tstrut & MAP\Tstrut & MRR\Tstrut \\
\midrule
1\Tstrut & 0.651\Tstrut & 0.794\Tstrut & 0.716\Tstrut & 0.842\Tstrut  \\
2 & 0.653 &	0.797 &	0.721 &	0.850 \\
3 & 0.698 &	0.830 &	0.800 &	0.908 \\
\textbf{4} & \textbf{0.734} &	\textbf{0.853} &	\textbf{0.863} &	\textbf{0.945} \\    
5 & 0.700 &	0.827 &	0.803 &	0.906 \\
6 & 0.457 &	0.606 &	0.467 &	0.621 \\
\bottomrule
\end{tabular}
\caption{
Model performance with original (5.5B) version of ELMo (top scores marked in bold) as the number of hop increases.
}
\label{table:performance_hop}
\end{table}
\begin{table}[t]
\small
\centering
\begin{tabular}{C{0.27\columnwidth}C{0.11\columnwidth}C{0.11\columnwidth}C{0.11\columnwidth}C{0.11\columnwidth}}
\toprule
\multicolumn{1}{c}{\multirow{2}{*}{\textbf{\# hop}\Tstrut}} & \multicolumn{2}{c}{dev}\Tstrut & \multicolumn{2}{c}{train}\Tstrut \\
\cmidrule{2-5}
\multicolumn{1}{c}{} & MAP\Tstrut & MRR\Tstrut & MAP\Tstrut & MRR\Tstrut \\
\midrule
1\Tstrut & 0.648\Tstrut & 0.790\Tstrut & 0.708\Tstrut & 0.842\Tstrut \\
2 & 0.655 &	0.801 &	0.720 &	0.853 \\
3 & 0.681 &	0.816 &	0.768 &	0.886 \\
4 & 0.706 &	0.834 &	0.796 &	0.906 \\    
\textbf{5} & \textbf{0.716} &	\textbf{0.841} &	\textbf{0.813} &	\textbf{0.916} \\
6 & 0.441 &	0.596 &	0.452 &	0.600 \\
\bottomrule
\end{tabular}
\caption{
Model performance with small version of ELMo (top scores marked in bold) as the number of hop increases.
}
\label{table:performance_hop_small}
\end{table}

\begin{figure*}[t]
\centering
\subfigure[Type-1]{\label{fig:topology_type1}\includegraphics[width=0.5\columnwidth]{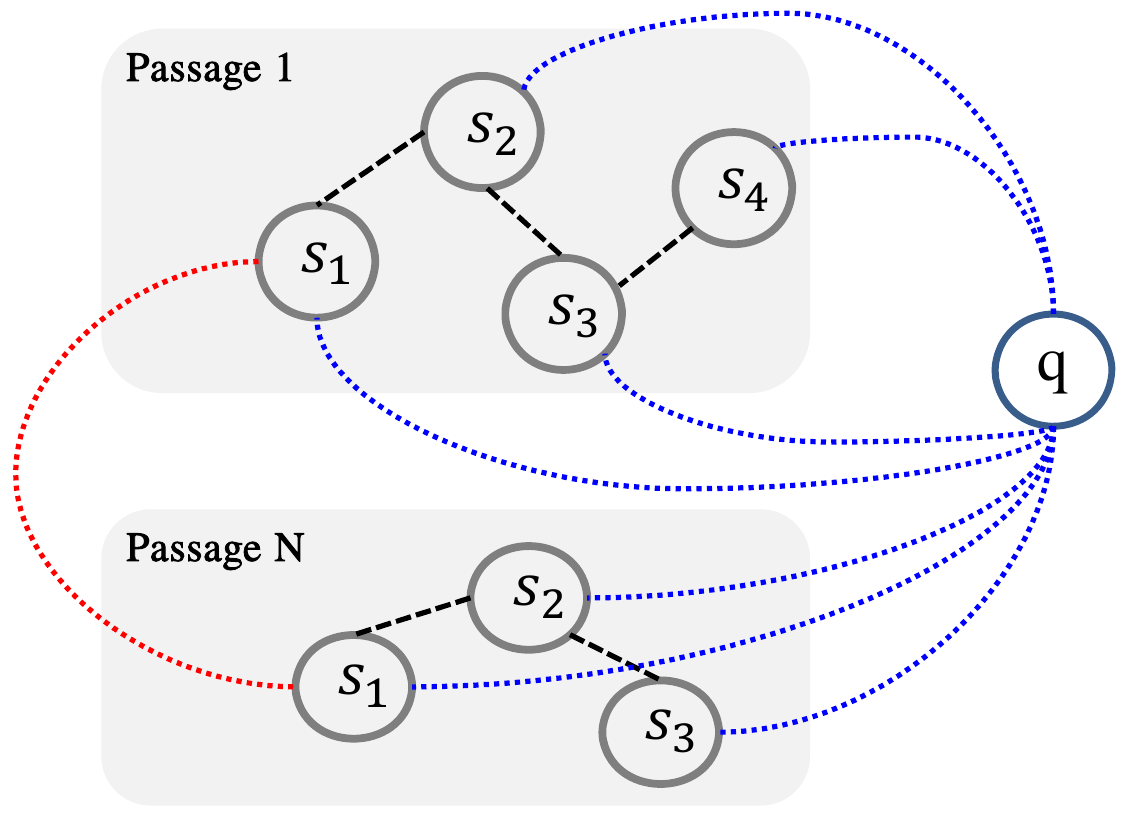}}
\qquad
\qquad
\subfigure[Type-2]{\label{fig:topology_type2}\includegraphics[width=0.5\columnwidth]{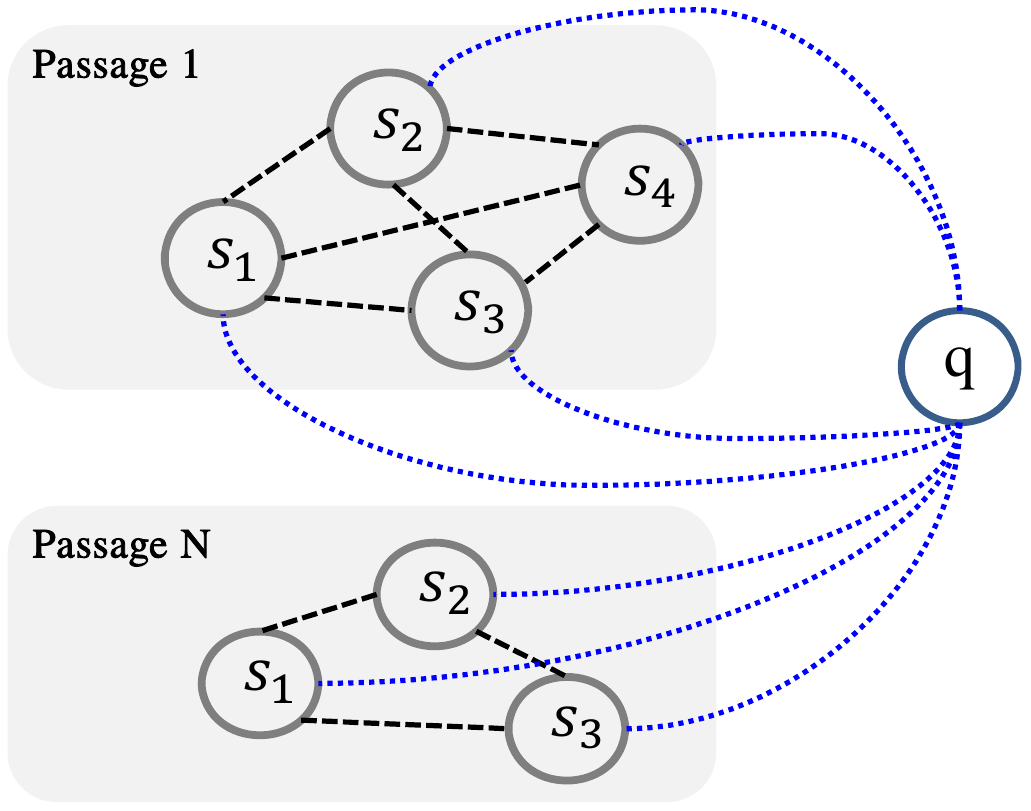}}
\qquad
\qquad
\subfigure[Type-3]{\label{fig:topology_type3}\includegraphics[width=0.5\columnwidth]{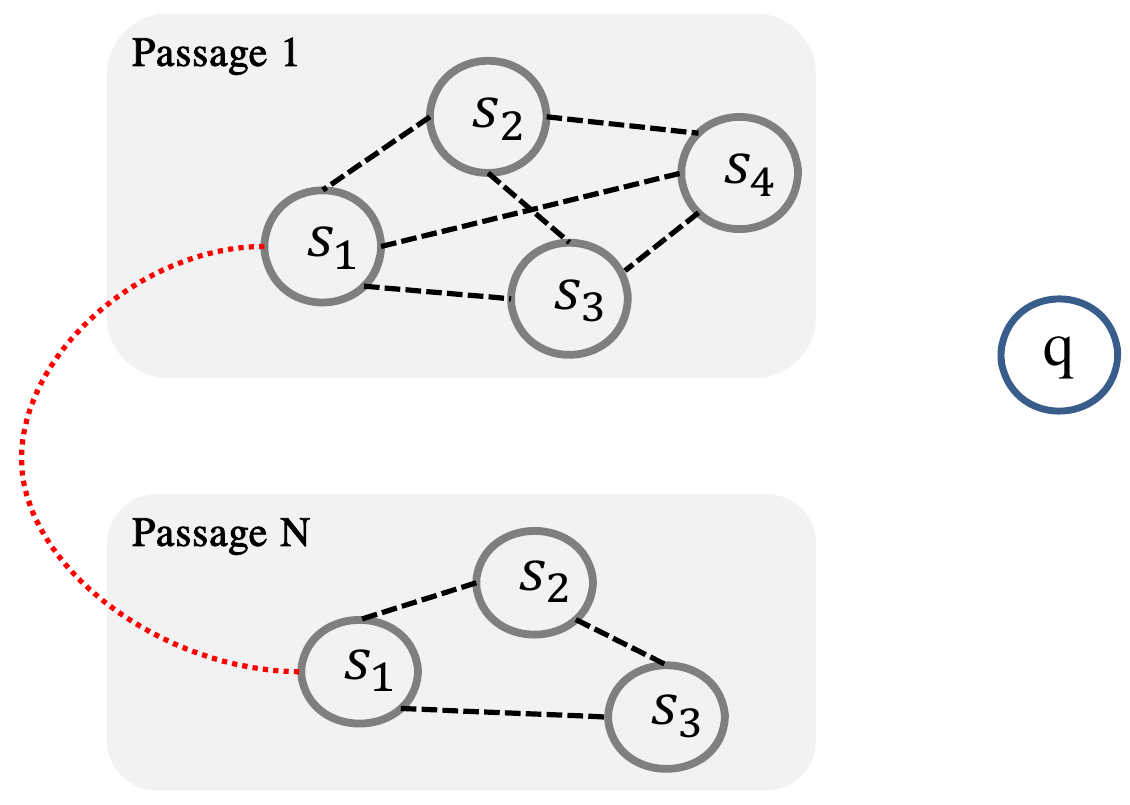}}
\caption{
Different typologies for the graph. Type-1 reduce connection within the passage, type-2 remove connection between the passages and type-3 remove connection between each sentence node and the question node.
}
\label{fig:topology_analysis}
\end{figure*}

\subsection{Impact of Various Graph Topologies}
\label{ssec:topology_analysis}
The topology of the graph determines the path by which information flows and is aggregated. To see the quantitative contributions of each connection in the graph, we perform ablation experiments as follows:
\begin{itemize}
    \item 
    \textbf{Type-1}: We reduce the connections between sentences within the same passage. Only the previous and next sentences are connected to their neighbor sentence (see figure~\ref{fig:topology_type1}).
    \item
    \textbf{Type-2}: We remove the connections between the passages, which reveals the contribution of the information flow among independent passages (see figure~\ref{fig:topology_type2}).
    \item
    \textbf{Type-3}: We remove the connections between each sentence node and the question node (see figure~\ref{fig:topology_type3}).
\end{itemize}

Figure~\ref{fig:topology_analysis} illustrates different types of connection strategies, and Table~\ref{table:topology_analysis} shows their corresponding performances. To reach the best performance, we conduct experiments multiple times by changing the number of hops in the model from 1 to 6 for each case (Type-1 to Type-3). 
From the experiment, hop-4 is selected as the best-performing hyper-parameter.
However, all the model variations undergo performance degradation compared to the original topology (\textbf{PS}-\textit{rnn-elmo-s}).

\begin{table}[t]
\small
\centering
\begin{tabular}{L{0.35\columnwidth}C{0.09\columnwidth}C{0.09\columnwidth}C{0.09\columnwidth}C{0.09\columnwidth}}
\toprule
\multicolumn{1}{l}{\multirow{2}{*}{\textbf{Model}\Tstrut}} &  \multicolumn{2}{c}{dev}\Tstrut & \multicolumn{2}{c}{train}\Tstrut \\
\cmidrule{2-5}
\multicolumn{1}{c}{} & MAP\Tstrut & MRR\Tstrut & MAP\Tstrut & MRR\Tstrut \\
\midrule
\textbf{PS}-\textit{rnn-elmo-s}\Tstrut & \textbf{0.716}\Tstrut & \textbf{0.841}\Tstrut & \textbf{0.813}\Tstrut & \textbf{0.916}\Tstrut \\
\midrule
\textbf{Type-1}~(\textit{rnn-elmo-s})\Tstrut & 0.694\Tstrut & 0.834\Tstrut & 0.807\Tstrut & 0.915\Tstrut \\
\textbf{Type-2}~(\textit{rnn-elmo-s}) & 0.705 & 0.836 & 0.792 & 0.903 \\
\textbf{Type-3}~(\textit{rnn-elmo-s}) & 0.658 & 0.796 & 0.729 & 0.857 \\
\bottomrule
\end{tabular}
\caption{
Model performance with different typologies. 
The connection strategies between nodes for each type are illustrated in figure~\ref{fig:topology_analysis}.
}
\label{table:topology_analysis}
\end{table}

\subsection{Impact of Node Representation}
\label{ssec:node_analysis}
To see the effectiveness of the various approaches for computing sentence representation, we investigate combinations of well-studied methods.
\subsubsection{Word Representation}
Vector representations of the words in each sentence are computed from the original version of the ELMo model (-elmo), a small version of the ELMo model (-elmo-s), BERT~\cite{devlin2019bert} model (-bert), or mapped to GloVe word embedding (-glove).

\subsubsection{Node Representation}
Each node representation is computed by employing three general methods for encoding the sequence of word representations as follows:

\begin{itemize}
    \item 
    We employ an RNN model (-rnn) to encode sequential information in the sentence. The final representation of the RNN's hidden status is considered as a node representation (see  equation~(\ref{eq:node_representation})).
    \item
    We apply a pooling method (-avg) that averages all the word representations in the sentence to compute the node representation as follows:  $\textbf{N}^Q{=}\,\text{average}(\textbf{Q})$, $\textbf{N}^S{=}\,\text{average}(\textbf{S})$. These new representations-$\textbf{N}^Q$ and $\textbf{N}^S$-are substituted for the node representations in equation~(\ref{eq:node_representation}).
    \item
    We adopt the pretrained universal sentence-encoding method (-USD\_T), which is based on the recently developed transformer model~\cite{cer2018universal}. This model computes sentence representation directly from the sequence of words in any text.
\end{itemize}

\begin{figure*}[t]
\centering
\subfigure[exact match]{\label{fig:em}\includegraphics[width=0.44\columnwidth, frame]{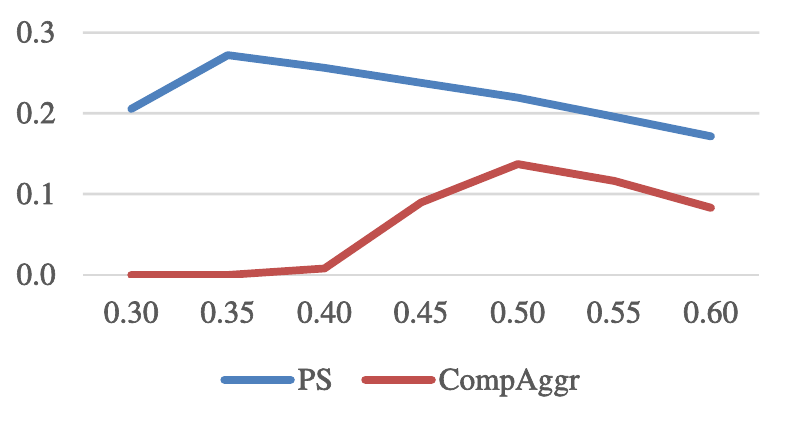}}
\qquad
\subfigure[recall]{\label{fig:recall}\includegraphics[width=0.44\columnwidth, frame]{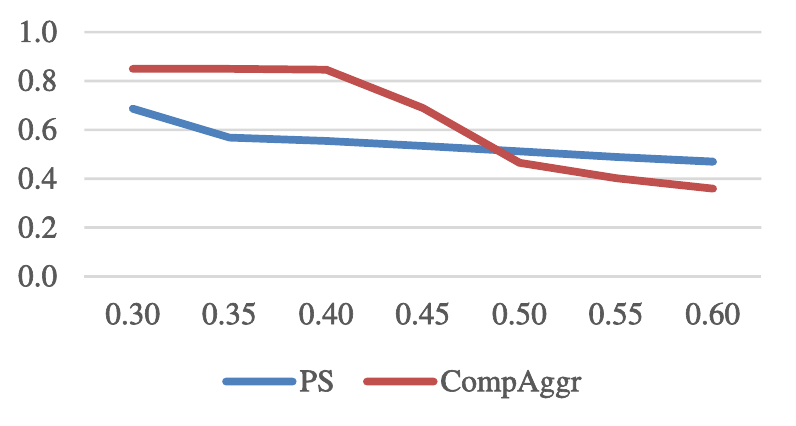}}
\qquad
\subfigure[precision]{\label{fig:precision}\includegraphics[width=0.44\columnwidth, frame]{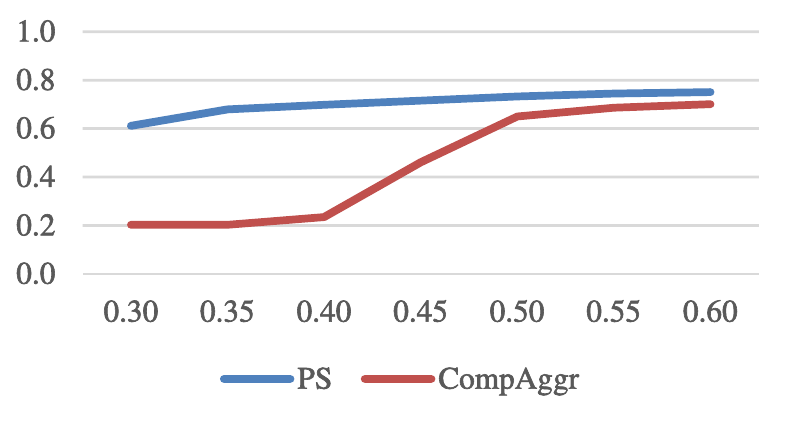}}
\qquad
\subfigure[f1]{\label{fig:f1}\includegraphics[width=0.44\columnwidth, frame]{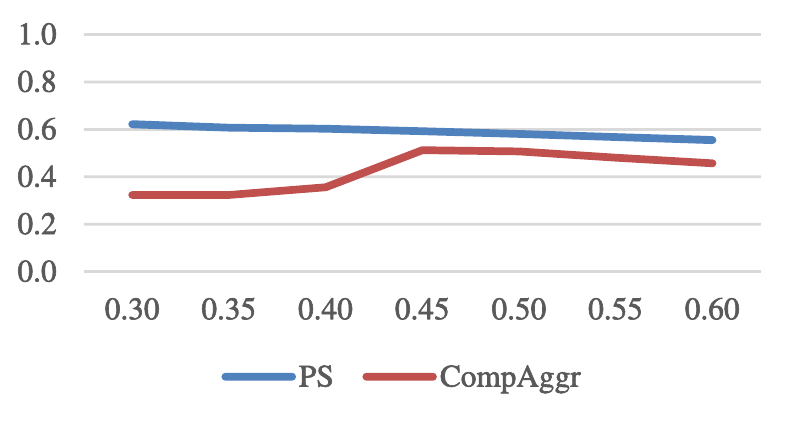}}
\caption{
Model performance with various measure. The x-axis shows a threshold value that is used for determining the label of the question-supporting sentence pair by the confidence score. 
}
\label{fig:evaluation}
\end{figure*}

\begin{table}[t]
\small
\centering
\begin{tabular}{L{0.31\columnwidth}C{0.10\columnwidth}C{0.10\columnwidth}C{0.10\columnwidth}C{0.10\columnwidth}}
\toprule
\multicolumn{1}{l}{\multirow{2}{*}{\textbf{Model}\Tstrut}} &  \multicolumn{2}{c}{dev}\Tstrut & \multicolumn{2}{c}{train}\Tstrut \\
\cmidrule{2-5}
\multicolumn{1}{c}{} & MAP\Tstrut & MRR\Tstrut & MAP\Tstrut & MRR\Tstrut \\
\midrule
\textbf{PS}-\textit{USD\_T}\Tstrut  & 0.651\Tstrut & 0.795\Tstrut & 0.693\Tstrut & 0.830\Tstrut \\
\midrule
\textbf{PS}-\textit{avg-glove}\Tstrut  & 0.617\Tstrut & 0.753\Tstrut & 0.876\Tstrut & 0.945\Tstrut \\
\textbf{PS}-\textit{avg-elmo-s} & 0.471 & 0.611 & 0.483 & 0.625 \\
\midrule
\textbf{PS}-\textit{rnn-glove}\Tstrut  & 0.700\Tstrut & 0.822\Tstrut & \textbf{0.919}\Tstrut & \textbf{0.971}\Tstrut \\
\textbf{PS}-\textit{rnn-elmo-s} & 0.716 & 0.841 & 0.813 & 0.916 \\
\textbf{PS}-\textit{rnn-elmo} & \textbf{0.734} & \textbf{0.853} & 0.863 & 0.945 \\
\textbf{PS}-\textit{rnn-bert} & 0.667 & 0.806 & 0.708 & 0.841 \\
\bottomrule
\end{tabular}
\caption{
Model performance with the different method for computing node representation. 
}
\label{table:node_representation}
\end{table}

Table~\ref{table:node_representation} depicts the model performance with different node representation methods. In all cases, the RNN encoding skims (-rnn) performs better than that of the average pooling (-avg). 
Interestingly, average pooling with ELMo representation (\textbf{PS}-\textit{avg-elmo-s}) performs worse than in the GloVe representation (\textbf{PS}-\textit{avg-glove}) case.
From this result, we find that averaging ELMo does not produce proper node representations.
For the \textbf{PS}-\textit{rnn-bert} case, we do not fine-tune the BERT model and only use its computing word representation. We expect there exists a possibility to enhance model performance by fine-tuning the BERT with the end-to-end training process.

\section{Discussion}
\label{sec:discussion}
In this study, we focus on a model that can detect \textit{supporting sentences} to answer a question.
We do not consider competing against the full QA systems, i.e., machine reading QA (MRQA) models, which are jointly trained with two objectives, ``\textit{extracting answer span}" and ``\textit{detecting supporting sentence}."
Note that we do not use the exact ``answer span" information when detecting the \textit{\textit{supporting sentences}}. We think ``answer-span" supervision allows the model to track the \textit{supporting sentences} from simple word matching.
Therefore, our investigations are focused on evaluating and analyzing the effectiveness of the proposed graph neural network-based model for classifying \textit{supporting sentences} compared to the well-known answer-selection QA models.
To evaluate the performance of the proposed model from a different perspective, we adopt other traditional measures for the QA system (i.e., precision, recall, and f1), and evaluate our methods. 
These measures require a specific label (\textit{true} or \textit{false}) for each pair of data (the question and the supporting sentence candidate).
As our model computes the confidence scores for each pair of data, we give a \textit{true}-label when the confidence scores are greater than a predefined threshold value (otherwise, we give the pair a \textit{false}-label).
Figure~\ref{fig:evaluation} shows the model performances (\textbf{PS}-\textit{rnn-elmo} vs \textbf{CompAggr}) in regards to the variation of the threshold value (0.3 to 0.6).

In future research directions, we will investigate the best way to combine our proposed model with existing MRQA algorithms to build a full QA system. 
It would also be possible to link the current graph to another graph (i.e., knowledge graph) to engage external knowledge information in the question-answering system.
We also hope that our work inspires future works aiming to perform multihop reasoning on the free-form text.

\section{Conclusion}
\label{sec:conclusion}
In this paper, we propose a graph neural network that finds the sentences crucial for answering a question.
The experiments demonstrate that the model correctly classifies \textit{supporting sentences} by iteratively propagating the necessary information through its novel architecture.
We believe that our approach will play an important role in building a QA pipeline in combination with other MRQA models trained in an end-to-end manner.

\section{Acknowledgements}
K. Jung is with ASRI, Seoul National University, Korea. This work was supported by the Ministry of Trade, Industry \& Energy (MOTIE, Korea) under Industrial Technology Innovation Program (No.10073144) and by the NRF grant funded by the Korea government (MSIT) (NRF2016M3C4A7952587).

\section{Bibliographical References}
\label{sec:references}

\bibliographystyle{lrec}
\bibliography{lrec20}


\end{document}